\documentclass[10pt,twocolumn,letterpaper]{article}

\usepackage{cvpr}
\usepackage{times}
\usepackage{epsfig}
\usepackage{graphicx}
\usepackage{amsmath}
\usepackage{amssymb}

\cvprfinalcopy 
\usepackage[pagebackref=true,breaklinks=true,letterpaper=true,colorlinks,bookmarks=false]{hyperref}

\usepackage{epsfig}
\usepackage{graphicx}
\usepackage{amsmath}
\usepackage{amssymb}
\usepackage{subfigure}
\usepackage{amsfonts}
\usepackage{setspace}
\usepackage{colortbl}
\usepackage{morefloats}
\newcommand{\RE}{\textcolor{black}}
\newcommand{\RV}{\textcolor{black}}

\def\cvprPaperID{3070} 

\ifcvprfinal\fi
\begin{document}

\title{Rank of Experts: Detection Network Ensemble}



\author{Seung-Hwan~Bae$^{1}$, Youngwan  Lee$^{2}$,  Youngjoo  Jo$^{2}$, Yuseok   Bae$^{2}$, Joong-won  Hwang$^{2}$\\
$^{1}$Computer Vision Laboratory, Incheon National University, South Korea\\
$^{2}$ Electronics and Telecommunications Research Institute, Daejeon, South Korea\\
{\tt\small shbae@inu.ac.kr, \{yw.lee,run.youngjoo,baeys,jwhwang\}@etri.re.kr}
}

\maketitle

\begin{abstract}
The recent advances of convolutional detectors show \RV{impressive performance improvement} for large scale object detection. However, in general, the detection performance  usually decreases as the object classes to be detected increases, and it is a practically challenging  \RV{problem} to train a dominant model for all classes due to the limitations of detection models and  \RV{datasets}. In  most cases, therefore, there \RV{are} distinct performance \RE{differences} of the modern convolutional detectors for each object class \RV{detection}.  In this paper, \RV{in order to build an ensemble detector for large scale object detection}, we present a conceptually simple but very effective  class-wise \RV{ensemble detection which is named as Rank of Experts. We first decompose an intractable problem of finding the best detections for all object classes into small subproblems of finding the best ones for each object class. We then solve the detection problem by ranking detectors in order of the average precision rate for each class, and then aggregate the responses of the top ranked detectors (\ie experts) for class-wise ensemble detection.} The main benefit of our method is easy to implement and does not require any \RE{joint training of experts} for ensemble. Based on the proposed \RE{Rank of Experts}, we won the 2nd place  \RV{in the ILSVRC 2017 object detection competition}.
\end{abstract}

\section{Introduction}
\label{Intro}
An object detection problem is to predict object \RV{hypotheses} of one or more classes given an image. \RV{In general, an object detection process consists of  feature extraction, region search and object classification, and classical object detectors are constructed by combing them.} However, the recent advances of deep learning \RV{change} the detection paradigm. In particular, the region-based convolutional detection approach \cite{SzegedyTE_NIPS13, Girshick15_ICCV15, SermanetEZMFL_CORR13} \RV{enables} us to detect \RV{large classes of objects} with a single convolutional neural network \RE{(CNN)}.

In the recent years, in order to detect \RV{large classes of objects},  deep \RE{CNNs} \cite{SimonyanZ14a, HeZRS_ECCV16, SzegedyIVA_AAAI17} with high classification rate trained on the ImageNet dataset \cite{ILSVRC15} are used for extracting object features and combined with several meta-architectures (\ie detectors) such as Faster RCNN \RE{(FRCN)} \cite{RenHGS15_NIPS15}, SSD \cite{LiuAESRFB_ECCV16}, DSSD \cite{FuLRTB_Corr17} and R-FCN\cite{DaiLHS_CORR16}. However, it is still challenging to train a single dominant model for all object classes due to the limitations of models and datasets. Also, in general, the detection performance is degraded \RV{as object classes increase}.

\begin{figure}[!tbp]
\includegraphics[width=1.00\linewidth]{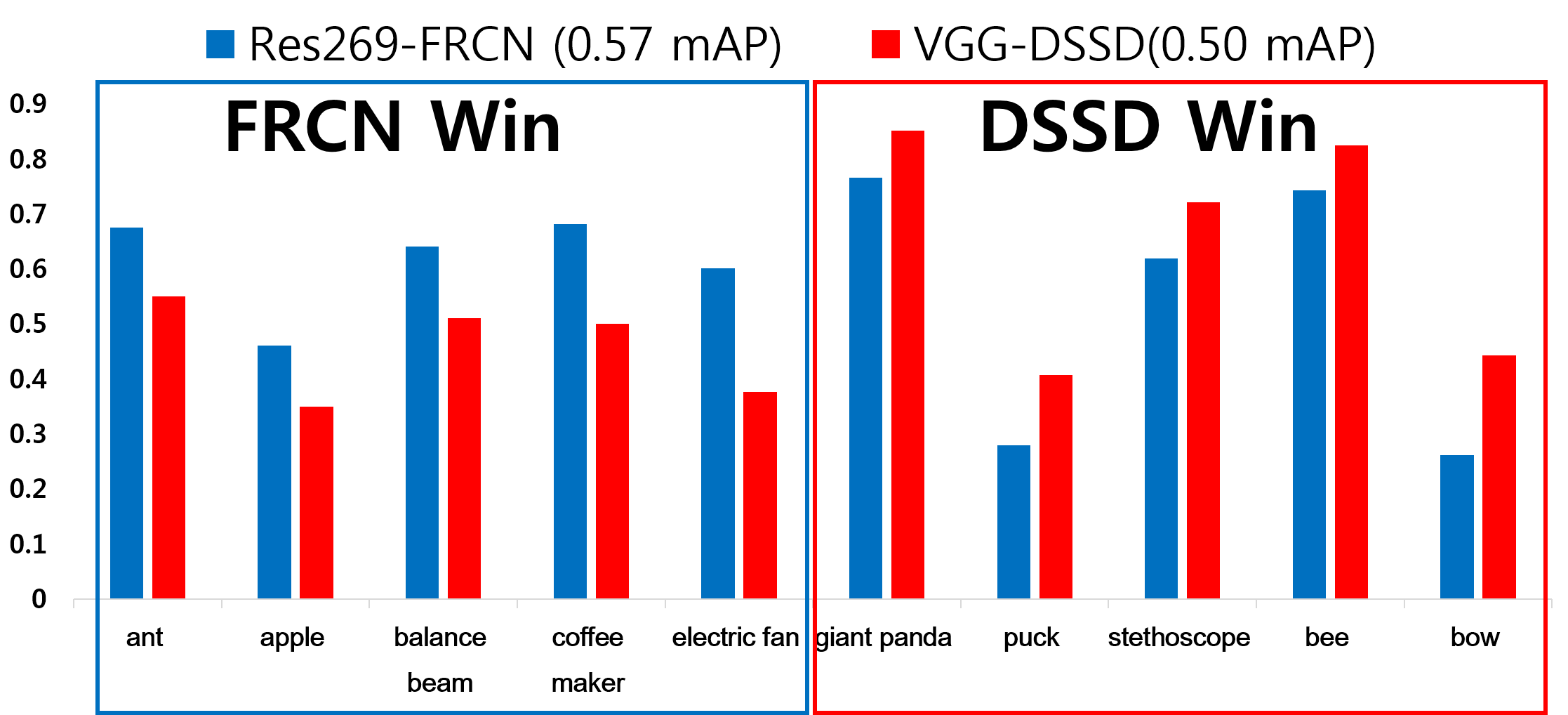}
\caption{\RE{Performance comparison between FRCN \cite{RenHGS15_NIPS15} using ResNet \cite{HeZRS_ECCV16} with 269 layers  and DSSD \cite{HeZRS_ECCV16} using VGG \cite{SimonyanZ14a} with 16 layers on the  ILSVRC 2017 val2 set. Even though the FRCN has the better mAP, the DSSD shows the better average precision rate for some \RV{object} classes.}}
\vspace{-5pt}
\label{fig:comp_frcn_dssd}
\end{figure}

To resolve this problem, we therefore propose an \RE{ensemble detection} \RV{in order to combine} different types of detectors for large \RV{scale} object detection. In this paper we named the proposed \RE{ensemble detection} as \RV{\emph{Rank of Experts}} since it can determine\ \RE{best} detectors (\textit{or} experts) for each class object detection. \RE{The motivation of} our \RE{ensemble detection} is that detectors with different network structures  have different detection rate for object classes \RE{in general}. In addition, we presume that a detector with the low mean average precision (mAP) can detect object instances of some classes better than a detector with high mAP. \RE{To support our assumption}, we compare the average precision between the FRCN and DSSD. As shown in Fig. \ref{fig:comp_frcn_dssd}, for some object classes the DSSD shows the \RE{better detection rates} even though the \RE{FRCN}  has the higher mAP than \RE{the DSSD}.

\RE{Based on the assumption}, in this study \RE{our main idea} is to develop the class-wise \RE{ensemble detection} which can determine the best experts and combine their detections effectively for detecting instances of the \RE{specific object class}. To this end, we first generate \RE{a pool of detectors}. In order to enhance the diversity of the detectors, we trained a set of detectors with \RV{various} feature extractors and meta-architectures. Once the \RV{detector pool} is generated, we generate a ranking matrix which \RE{represents} the ranking of the detectors according to their average precision on each class. We then combine responses of top rankers of the pool for detecting the instances of the corresponding class.

Although our method is rather simple, there are obvious benefits of the proposed \RE{ensemble detection}. Most of all, it can be easily implemented and does not require any \RE{extra joint training of detectors} for ensemble. In addition, in most cases, the performance of the ensemble model is better than that of best single detector \RV{since it can selectively combine the good detection results of multiple detectors.}

Recently, the 1st place winners at ILSVRC \cite{ILSVRC15} \& MSCOCO \cite{LinMBHPRDZ_corr14} 2015/2016 competitions improve the  performance further using the model ensemble. MSRA \cite{HeZRS_CVPR16} \RV{generates} a union set of region proposals of models and the set is processed by an ensemble of per-region classifiers. \RV{However, the computational complexity increases significantly as the number of models increases since region proposals of whole models should be processed by each individual detector.} For more efficient network ensemble,  Google \cite{HuangRSZKFFWSG_Corr16} \RV{combines} models with complementary strength only. To this end, they greedily select models with the highest mAP on a validation set, but prune away a model which shows high similarity for the category AP with the selected ones. However, mAP \RV{is} an indirect metric to select models for \RV{class-wise} ensemble \RV{since high mAP does not always ensure the superiority on class-wise performance as shown in Fig. \ref{fig:comp_frcn_dssd}}.
On the other hand,  we use the more  \RE{direct and meaningful average precision  metric for class-wise ensemble}. As a result, we can select and combine the best \RE{experts for each category detection}, and it can improve the class-wise performance.

To prove the effects of our \RE{ensemble detection}, we generate a \RV{detector} pool consisting of different types of detectors. Each detector was implemented based on recently developed feature extractors (ResNet101/152/269 \cite{HeZRS_CVPR16}, VGGNet \cite{SimonyanZ14a} and WRI(wide-residual-inception)Net \cite{LeeKPCK_Corr17}) and \RE{meta-architectures} (\RE{FRCN} \cite{RenHGS15_NIPS15}, SSD \cite{LiuAESRFB_ECCV16} and DSSD \cite{FuLRTB_Corr17}). By combining them based on our ensemble method, we greatly improve the mAP up to about 4\% and 5\% on \RE{PASCAL VOC 2007/2012 (07/12)  and ILSVRC 2017 datasets, respectively, as \RV{shown} in Table \ref{TABLE:Comp_PASCAL} and \ref{TABLE:Comp_ILSVRC}}. Remarkably, we won the 2nd place \RV{in the ILSVRC 2017  detection competition}.

To summarize, our main contributions are as follows: (1) \RE{proposition of} \RV{a new ensemble method} to combine different detectors effectively \RE{without extra joint training} and improve the class-wise performance directly, (2) \RV{extensive implementation of modern convolutional detectors with \RV{various} feature extractors and meta-architectures to generate a detector pool with high diversity}, (3) \RE{thorough comparison of} the proposed methods with recent detection and ensemble methods, and the details of our approach for winning the object detection challenge.

\section{Related Work}
We discuss previous works on deep networks and network ensemble for  detection, which are  related to our work.

\textbf{Detection network:}
Since Krizhevsky et al. \cite{KrizhevskySH_NIPS12} \RE{achieved remarkable} \RV{progress} using CNN in object classification, a series of detection methods based on CNNs have been flourished. In RCNN \cite{GirshickDDM_CVPR14}, CNN is used to extract feature maps from warped regions of interest (\RV{RoI}) \RV{generated} by selective search \cite{UijlingsSGS_IJCV13}. Spatial pyramid pooling network \cite{kaiming14ECCV} improved RCNN by sharing feature maps of images instead of extracting a CNN feature for each \RV{RoI}.

For \RV{end-to-end training}, Fast \RE{RCNN} \cite{Girshick15_ICCV15} introduces a single-stage training algorithm which can jointly learn to classify object proposals and refine their locations. In addition, \RE{Faster RCNN} \cite{RenHGS15_NIPS15} further improves the speed and accuracy by embedding \RV{a} region proposal network  into the Fast RCNN.

In addition, \RE{many unified detection frameworks} such as YOLO \cite{RedmonDGF_CVPR16}, SSD \cite{LiuAESRFB_ECCV16}, and R-FCN \cite{DaiLHS_CORR16} trained with \RV{a} multi-task loss function for object classification and regression have been developed for the better speed. \RE{On the other hand, feature pyramid network \cite{LinDGHHB_Corr16} improves the accuracy and reduces computational complexity by constructing feature pyramid  instead of image pyramid.} \RE{Deformable} convolutional network \cite{jifeng_iccv17} \RE{proposes} deformable convolution to enhance the robustness of the CNN against the geometric transformation. 
\RE{Note that this paper mainly focuses on combining the modern detectors effectively for better detection rather than improving their performance.}

\textbf{Network ensemble:} 
The idea of ensemble learning, which combines multiple \RV{models}, has been widely used due to  two \RV{important} benefits. The first one is that it can usually enhance \RE{the} generalization performance. In fact, even \RV{models} with similar performance may show different generalization ability. By combining several \RV{models} with majority voting \cite{HansenS_PAMI90} for the classification or weighted averaging  \cite{FREUND1995256, Opitz99popularensemble, Perrone93whennetworks} for regression, the overall risk of making a particularly poor selection can be reduced  \cite{Polikar_CSM06}.

\RE{When creating an ensemble system,  increasing the diversity of the \RV{models}  can reduce the generalization error  \cite{Polikar_CSM06, KroghV_NIPS94}} in general. Therefore, it is important  to make \RV{models} produce different errors  \RV{to represent different regions of input space}. For achieving this, \RV{training models with data resampling} \cite{HansenS_PAMI90, KroghV_NIPS94, JacobsJNH_NC91, JordanJ_NC94} is one of the most popular methods. \RE{Training CNNs with different initial states and parameters  or the order of mini-batches can also generate different outputs}. AlexNet \cite{KrizhevskySH_NIPS12} \RV{increases} the diversity  by training the same CNNs with different initial parameters.  VGGNet \cite{SimonyanZ14a} \RV{selects} two best performing models among five models with distinct architecture. GoogleNet \cite{CVPR15_Inception} \RV{achieves the diversity using the resampling.}

Moreover, \RV{the other benefit is that} ensemble learning can handle  a complex  task based on  \RV{a} divide-and-conquer approach. It divides the task into \RV{small} subproblems and solve them with multiple learners. Mixture of experts \cite{JacobsJNH_NC91, Noam_1701}
that controls the activation of  experts with \RV{a gate function} on a given subtask is representative of this approach.

\RE{For combining models, \RV{the methods} \cite{Perrone93whennetworks, KroghV_NIPS94, JordanJ_NC94, ZhouWT_AI02} using the weighted averages  of \RV{network outputs} \RV{are} most well-known.  A new layer \cite{ParkHBB_WACV16} or gate networks \cite{LasotaLTT_ACIDS14, Noam_1701, David_ICLRW17} are trained to select appropriate models for a given task. Recently, He et al. \cite{HeZRS_CVPR16} \RV{collect} region proposals of several models and  forwards them to \RV{each RCNN}. Huang et al. \cite{HuangRSZKFFWSG_Corr16} select dissimilar models \RV{by using} the cosine similarity.} 

\RV{In this work, for improving the generality and diversity we build a detector pool with modern detectors and effectively aggregate their outputs \RV{by using} our Rank of Experts.}

\section{Detector Pool}
\label{Sec:Pool}

To enhance the diversity of our ensemble \RV{detection}, we generate a detector pool which contains convolutional detectors with \RV{various} feature extractors and meta-architectures. In particular, we have implemented \RV{several} modern convolutional detectors. We have combined them with the recent feature extractors, and trained them with the multi-task loss function \cite{Girshick15_ICCV15} \RV{for} object classification and localization. We provide the details of the implementations and evaluation results of each detector on PASCAL VOC \& ILSVRC datasets as  in Table \ref{TABLE:Comp_PASCAL} and \ref{TABLE:Comp_ILSVRC}, \RV{respectively}.

\subsection{Feature Extractor}
\RE{For feature extraction, we use the ResNet \cite{HeZRS_ECCV16}, VGGNet \cite{SimonyanZ14a} and WRINet \cite{LeeKPCK_Corr17}}. A ResNet  uses identify mapping with shortcut connection between feature maps  to propagate signals effectively in a deep network.  We implement several ResNets with different layers (101/152/269) by stacking different number of residual blocks. We also use the modified atrous VGG16 for reducing network parameters  \cite{LiuAESRFB_ECCV16}. Compared to original VGG16 \cite{SimonyanZ14a},  fully connected layers are replaced with convolutional layers or removed to reduce the computational cost and memory usage, but its accuracy is similar to \cite{SimonyanZ14a}. 

A WRINet \cite{LeeKPCK_Corr17} has a shallow and wide structure for fast  feature extraction with low memory. For accurate feature extraction for objects of various sizes, it is designed based on the combination of residual \cite{HeZRS_ECCV16} and inception \cite{CVPR15_Inception} modules. In specific, it consists of the basic residual blocks with two consecutive 3x3 convolutional layers and identity mapping, and inception-residual blocks with 1x1, 3x3 and 5x5 convolutional layers which are concatenated and followed by identity mapping. It has achieved higher accuracy  than the ResNet-101 with three times faster speed on \RV{the} KITTI and CIFAR-10/100 datasets.

Before combining extractors with \RV{meta-architecture}, it is possible to train them \RE{on} ILSVRC \cite{ILSVRC15} or other datasets to enhance their generalization ability. \RV{However, we use the pre-trained  ResNet \cite{ZengOYLXWLZYWZW_CORR16, tensorflow2015}, VGGNet \cite{LiuAESRFB_ECCV16} and WRINet \cite{LeeKPCK_Corr17} models only opened to the public because the focus of our work is not on improving the feature extractors.}
 
\subsection{Meta-Architecture}
\label{Sec:Meta}
Based on the modern convolutional detectors (\RE{FRCN} \cite{RenHGS15_NIPS15} and SSD \cite{LiuAESRFB_ECCV16}), we implement the original and extended versions using  feature map \RV{aggregation} \cite{KongYCS_CVPR16}, deconvolution \cite{FuLRTB_Corr17} and \RV{RoI} alignment \cite{HeGDG_CORR17}. The implemented detectors are \RE{listed} in Table \ref{TABLE:Comp_PASCAL} and \ref{TABLE:Comp_ILSVRC}.
%
%
All the detectors are designed based on a single CNN, and have in common the box (\textit{or} region) proposal, box classification and refinement components. 
%
The parameters of the components are trained with confidences and locations of predicted and ground truth boxes with a multi-task loss function through end-to-end learning, where the loss is a weighted sum of the classification $L_{cls}$ and localization $L_{reg}$ \RV{losses}  as $L =L_{cls}+ L_{reg} $ (see \cite{Girshick15_ICCV15} for more details of $L_{reg}$ and $L_{reg}$). 


\RE{For improving the performance, we implement FRCN-Type1 with the following modifications }: (1) we make the ratio of positive and negative samples in each min-batch \RE{equal}, and 
(2) \RV{we allow region proposals of the small sizes ( $\geq 0$) to be used for training in order to detect small objects}. These effects are proved in Table \ref{TABLE:Comp_Param}.


\begin{figure}[!tbp]
\includegraphics[width=1.00\linewidth]{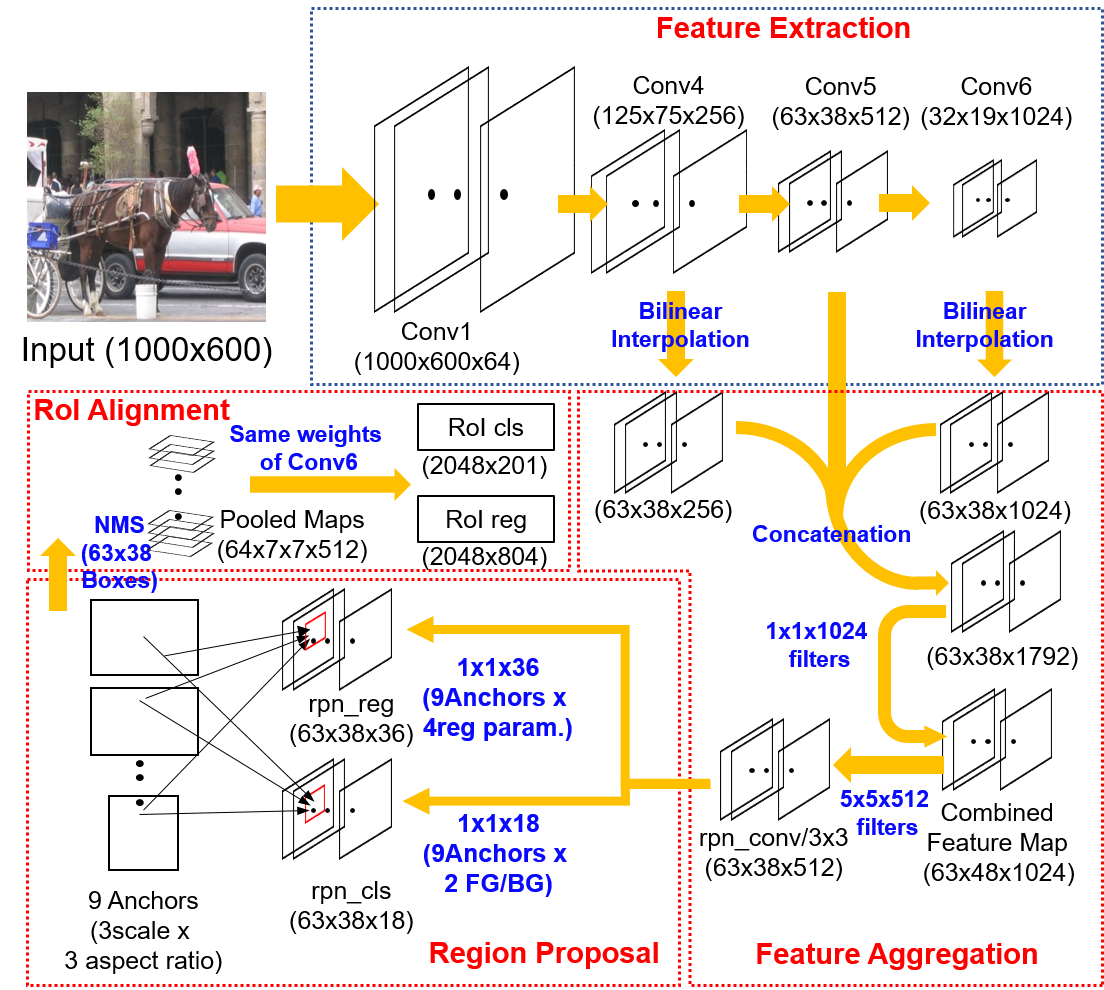}
\caption{\RV{The proposed FRCN-Type2 detector with feature aggregation and RoI Aligment.}}
\label{fig:Feature_comb}
\end{figure}

In FRCN, a feature map of the fixed size (\eg 7x7) is extracted by the RoIPooling operation from each proposal, and the feature map is fed to higher layers for classification and box regression. \RE{We also implement the variant of \RE{FRCN} by replacing the RoIPooling layer with the \RV{RoIAlignment} layer \cite{HeGDG_CORR17}, which uses bilinear interpolation instead of the hard quantization.} 
In addition, since exploiting multiple feature maps enhances the FRCN accuracy as shown \cite{KongYCS_CVPR16}, we implement {\RE{FRCN-Type2}} by aggregating multi-feature of several layers and use the aggregated feature for region/RoI proposal and classifications as shown in Fig. \ref{fig:Feature_comb}. However, our implementation is different from \cite{KongYCS_CVPR16}: we aggregate feature maps of different layers (Conv4/5/6) using bilinear interpolation \RV{to resize} \RV{them} instead of aggregating feature maps of layers (Conv1/3/5) using pooling and deconvolution. As a result, we can achieve the similar accuracy to  \cite {KongYCS_CVPR16} but improve the speed \RV{by avoiding the expensive pooling and deconvolution operations, and using features of the small sizes for aggregation.}

We \RV{also} implement SSD \cite{LiuAESRFB_ECCV16} and DSSD \cite{FuLRTB_Corr17}. DSSD \RV{improves} the SSD for the small object detection by merging encoded and decoded feature maps. Remarkably, as discussed in \cite{FuLRTB_Corr17, HuangRSZKFFWSG_Corr16}, SSD and DSSD show lower \RE{accuracies for the small objects} but higher \RE{those} for the large objects than \RE{FRCN}. Thus, we apply those detectors for \RE{ensemble detection} to improve the robustness against object scale changes.

\section{Rank of Experts}
\label{Ranking}
\RE{As mentioned, the main benefits of network ensemble are to improve the generalization performance and handle a difficult task using the divide-and-conquer approach. 
For large scale object detection, we also use the network ensemble for improving the generalization ability by combining multiple  different detectors  and dividing a difficult detection problem into small subproblems, where we reduce a intractable problem of finding the \RV{best detection responses} for  all the object classes to small subproblems of finding the best ones for each object class.}

\RE{In the most existing ensemble methods, a joint training between multiple networks is usually required. The weights  of the networks (\ie  beliefs in networks) are learned jointly by minimizing the generalize loss of an ensemble model \cite{JacobsJNH_NC91, ZhouWT_AI02,Polikar_CSM06}, where the loss is evaluated with the residual between desired outputs and the networks' combined outputs with weights.  A new layer is trained to control the amount of feature maps between the networks \cite{ParkHBB_WACV16}. To allocate  appropriate experts for a given task, \RV{they} \cite{JacobsJNH_NC91, JordanJ_NC94,HuangRSZKFFWSG_Corr16, LasotaLTT_ACIDS14} train an additional gating network together.}

\RE{However, we found that it is not effective to apply the traditional ensemble learning approach directly for large scale object detection. The main reason is that learning \RV{several deep networks (\eg ResNet and GoogleNet)} together is not practical because of huge GPU memory usage and expensive computational complexity. In addition, the diversity between networks is not much} \RV{when training several detectors using only data resampling  as done in \cite{JacobsJNH_NC91, JordanJ_NC94, KroghV_NIPS94, ZhouWT_AI02}.} 

\RE{Unlike the previous works, the proposed Rank of Experts is designed for ensemble \RV{appropriate for} large scale object detection. This allows us to combine deep convolutional detectors without the expensive joint training between them. By exploiting the proposed ensemble method, we can determine best convolutional detectors and combine their outputs for the specific object class detection. In the proposed object detection, we further improve their diversity by training detectors with \RV{various} feature extractors and meta-architectures as well as data resmapling and augmentation as discussed in  Sec. \ref{Sec:Pool} and \ref{Exp}. Our \RV{Rank of Experts} is written in Python. }

\subsection{Ensemble Detection Formulation}
Given an input image, a trained detector \RV{$T_{i}$} generates a set of detections $D_{i} = \left\{\mathbf{d}_{k}|k=1, ...,M_{i}\right\}$, where $M_{i}$ is the number of \RV{detection boxes} of \RV{$T_{i}$}. 
Each detection is represented as $\mathbf{d}_{k}=(\mathbf{p}_{k}, s_{k}, l_{k})$, where $\mathbf{p}_{k}$, $s_{k}$ and $l_{k}$ are the bounding box left top and right bottom position, confidence score and object class.  
When trained \RV{$N$} detectors are applied for an image, all detections of $N$ detectors can be defined as $\mathbb{D}= \bigcup_{i=1}^{N} D_{i}$. Similarly, we denote a set of ground truth boxes in the image as $\mathbb{G}$. Then, a detection problem can be formulated to find the optimal $\mathbb{D}$ by minimizing the loss function as 
\begin{equation}\label{eq:Ass_score1} 
\begin{array}{lc} 
\hat{\mathbb{D}} = \underset{\mathbb{D}}{\text{argmin}}~L_{cls}(\mathbb{D}, \mathbb{G}) + L_{reg}(\mathbb{D}, \mathbb{G})
\end{array}
\end{equation}
where the same classification $L_{cls}$ and regression loss  $L_{reg}$ are used as defined in \cite{Girshick15_ICCV15}. Since the ground truth is not available, the solving Eq. (\ref{eq:Ass_score1}) is not feasible in practice. To resolve this problem,  we divide the problem Eq. (\ref{eq:Ass_score1}) of finding ${\mathbb{D}}$ for all classes to small subproblems of finding $\mathbb{D}^{j}$ for class $j$. Then, the problem can solved in the following two phases: For each class $j$, we first evaluate the average precision of $N$ detectors and rank them according to its precision in a network ranking phase, and then collect outputs of the top ranked detectors to generate $\mathbb{D}^{j}$ in \RV{an} inference phase.  



\subsection{Network Ranking}
\RV{In order to rank detectors, we can exploit $K$-fold cross validation. We divide a training set into $K$ subsets. Then, we train a detector $T_{i}$ with $K-1$ sets and evaluate its average precision $AP^{k}(T_{i}^{j})$ for  object class $j$ with the \RV{held-out} set at round $k$. An average precision matrix for $N$ detectors for $C$ object classes  is denoted as    
}
\begin{equation}\label{eq:Rmatrix} 
\begin{array}{lc} 
 R =[r_{ji}]_{C \times N},~r_{ji} = \frac{1}{K} \sum_{k=1}^{K} AP^{k}(T_{i}^{j} ),
\end{array}
\end{equation}
%

Then, we generate a \RE{ranking} matrix $R^{*} = [r^{*}_{ji}]$ by sorting each of its rows in descending order, where $r^{*}_{ji}  \geq r^{*}_{ji+1},~i=1,...,N-1$. Here, it is \RV{worthy of notice} that we rank the detectors for each class with the AP measure instead of using the mAP for all classes. \RV{This makes detection results for each class decoupled}. As a result, we can select and combine best experts for each class regardless of the AP rates for other classes, and improve the class-wise performance directly.

\subsection{\RV{Ensemble Inference}}
Once a ranking matrix is generated, we select detectors with high \RV{APs} among trained detectors for each class. Let denote the number of selected ones for object class $j$ as $N^{j}$. To determine $N^{j}$ for class $j$, we evaluate a threshold \RE{$V^{j}$} by subtracting $\delta$ from $r^{*}_{j1}$ (\ie the maximum AP of $N$ detectors' APs for class $j$), and then select $T^{*}_{i}$ when $\RV{r^{*}_{ji}} \geq {V^{j}}$. 

For prediction, we  stack the outputs of selected detectors  as $\mathbb{D}^{j}= \bigcup_{i=1}^{N^{j}} D_{i}^{j}$, where  $D_{i}^{j}$ and $\mathbb{D}^{j}$ are the detection boxes of $T^{*}_{i}$ and \RV{an union set of  detections} for class $j$. We then apply soft non-maximum suppression \RV{(Soft-NMS)} \cite{BodlaSCD_CORR17} for $\mathbb{D}^{j}$, which decays $s_k$ of an overlapped box \RV{$\mathbf{d}_k$} with a box \RV{having} a maximum score rather than removing $\mathbf{d}_k$, because it shows the better AP scores than conventional NMS for several benchmark datasets. After the \RV{Soft-NMS}, we use the remaining boxes as \RV{final} detections of the object.

Note that before generating  $\mathbb{D}^{j}$ applying the \RV{Soft-NMS} for \RV{$\mathbb{D}_{i}^{j}$} could reduce the detection performance. The reason is that some false positive samples could not \RV{be} suppressed since the detections of other detectors, which can remove the false ones, \RV{are eliminated by the previous Soft-NMS}.

%


\section{Experiments}

\label{Exp}
\subsection{Implementation Details}
\label{Exp:Details}
\begin{figure}[!tbp]
\includegraphics[width=0.45\linewidth]{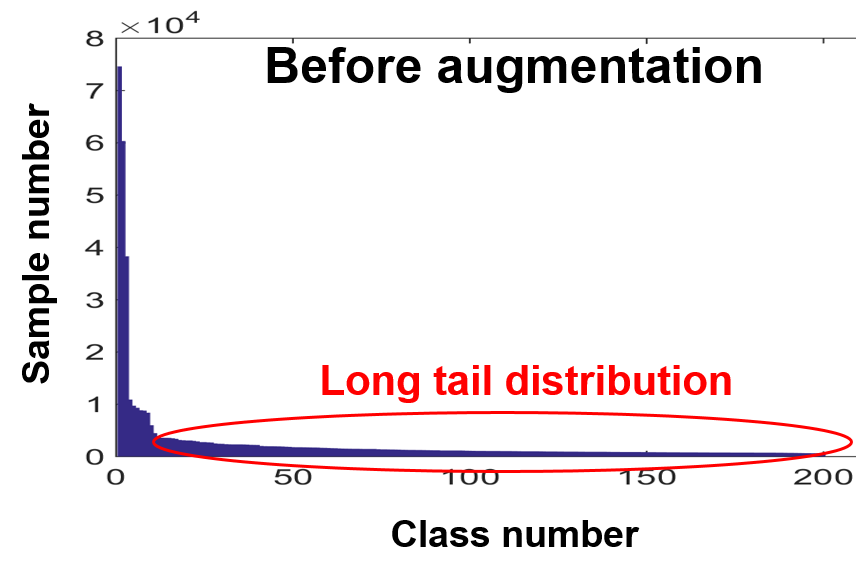}
\includegraphics[width=0.45\linewidth]{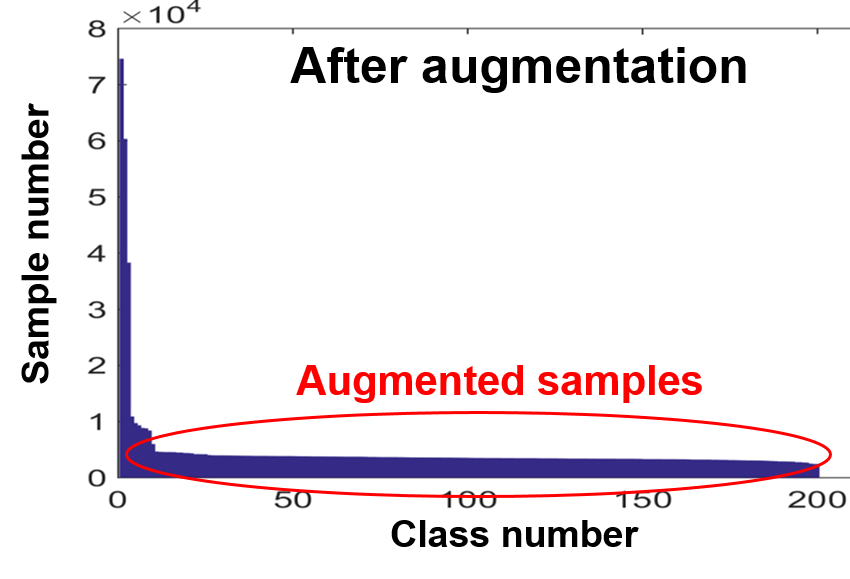}
\caption{{Class distribution before/after augmentation on the  ILSVRC 2017 training dataset}}
\label{fig:Samp_dist}
\end{figure}

\textbf{Data augmentation:}
For the ILSVRC 2017 detection dataset, we augment the training set to mitigate a data imbalanced problem between object classes. As shown in \RE{Fig.} \ref{fig:Samp_dist}, \RV{the training dataset} has long tail distribution, and fine-tuning models with it may degrade performance \cite{longtail}. 

\RE{The additional  benefit} is that we can generate different detectors by training them with/without the augmented samples as in Table \ref{TABLE:Comp_ILSVRC}. As a result, we can increase the diversity without using \RE{bagging which reduces a  training set}.

For data augmentation, we use  2D rotation, 2D translation, color histogram normalization, and noise addition. In 2D rotation, a whole image is rotated, but we restrict the rotation degree in a small range $(-6^{\circ}, ~6^{\circ})$ \RE{to maintain box annotation quality}. For 2D translation, we crop a (randomly selected) border of \RE{each} image. In FRCN the size of the feature map (\eg the last block layer of the ``conv4" block in ResNet to be extracted by RoIPooling) is 16 times smaller (by max pooling and 2 stride in ``conv1/3/4" block) than the input image size. For preventing the large translation of the feature map, we therefore cut $\{4, 8, 12\}$ pixels in the border after resizing. In the histogram normalization, histogram equalization, color casting, and color balancing are selectively applied. We also generate a noise image with additive white Gaussian noise with variance $[0.1, 0.001]$. 

We randomly select and apply one of those methods for each original image. The number of augmented images from \RE{each original image varies} 0 to 9. More samples are augmented when the sample number of the class is smaller. In our experiment, we generate 713,294 images with the 333,474 original images.  Figure \ref{fig:Samp_dist} shows the change of class distribution via data augmentation.

\textbf{Multi-scale testing:}
For more accurate detection, we also use multi-scale testing as used in \cite{kaiming14ECCV, Girshick15_ICCV15, HeZRS_CVPR16}. We apply the multi-scale testing only for the FRCN models and use four image scales $s \in \{400, 600, 800, 900\}$, where the $s$ is the length of an image's shorter side. We cap the longer size $l$ of an image's side at 1500 pixels for fitting the \RV{Res269-FRCN} in \RV{our} GPU memory during testing, and keep the image's aspect ratio when resizing. However, we have not performed multi-scale training (\cf \cite{kaiming14ECCV, Girshick15_ICCV15}). All the FRCN models are trained with $s=600$ and $l=1000$. Table $\ref{TABLE:Comp_ILSVRC}$ shows the detection results of the trained detectors at each scale. Since we trained those detectors with $s=600$, they have the best rates at $s=600$. \RV{We use detection results of the models for all the scales for ensemble detection.}

\begin{table}[tbp]
\caption{{The comparison of GPU memory usage with \RV{(w/t)} and
 without \RV{(w/o)} Caffe memory optimization. We test it on a IBM Minsky sever with 16280 (Mib) GPU memory. }}  
\vspace{-5pt}
{\footnotesize
\begin{center}
\renewcommand{\tabcolsep}{1.0mm}
\begin{tabular}{c|c|c}
\hline \hline
\textbf{Network} & \textbf{w/o optimization (Mib)}   & \textbf{w/t optimization (Mib)}  \\
\hline 
Res101-FRCN & 9731 & 5969 \\
Res152-FRCN &  12889 & 7771 \\
Res269-FRCN & Out of memory & 13581\\
\hline \hline 
\end{tabular}
\end{center}}
\vspace{-5pt}
\label{TABLE:Comp_GPU}
\end{table}

\begin{table}[tbp]
\vspace{-5pt}
\caption{{Comparison results of the Res-FRCN models with different $FG$ and $TR$ on PASCAL 2007 test set.}}  
\vspace{-0pt}
{\footnotesize
\begin{center}
\renewcommand{\tabcolsep}{0.7mm}
\begin{tabular}{c|c|c|c|c|c|c|c}
\hline \hline
\textbf{Network} & $\mathbf{FG}$ & $\mathbf{TR}$ & \textbf{mAP} & \textbf{Network} & $\mathbf{FG}$ & $\mathbf{TR}$ & \textbf{mAP}  \\
\hline 
Res101-FRCN & 0.25 & 0 & 76.31\% & Res101-FRCN & 0.25 & 16 & 76.75\%\\
Res101-FRCN & 0.5 & 0 & \color{black}76.82\% & Res101-FRCN & 0.5 & 16 & 76.52\%\\
\hline
Res152-FRCN & 0.25 & 0 & 75.71\% & Res152-FRCN & 0.25 & 16 & 76.76\%\\
Res152-FRCN & 0.5 & 0 & \color{black}76.92\% & Res152-FRCN & 0.5 & 16 & 75.99\%\\
\hline
Res269-FRCN & 0.25 & 0 & 77.31\% & Res269-FRCN & 0.25 & 16 & 76.75\%\\
Res269-FRCN & 0.5 & 0 & \color{black}77.40\% & Res269-FRCN & 0.5 & 16 & 76.90\%\\
\hline \hline 
\end{tabular}
\end{center}}
\label{TABLE:Comp_Param}
\vspace{-10pt}
\end{table}

\textbf{Batch normalization and memory optimization:}
We use optimized Caffe \cite{Caffe_TSN2016ECCV} for improving the speed of batch normalization.  We have inserted the batch normalization layer in the original \RE{FRCN} code \cite{RenHGS15_NIPS15}. For reducing GPU memory usage during network training and testing, we also apply the memory optimization code  \cite{Caffe_TSN2016ECCV} to \cite{RenHGS15_NIPS15}. It can reduce the memory \RV{usage} by reusing the memory \RV{allocated} by previous blobs in a network. Table \ref{TABLE:Comp_GPU} compares the memory usages when training different Res-FRCN models with a mini-batch with 128 samples.

\subsection{Performance Evaluation}
\textbf{Experimental setup:} We compare our \RV{Rank of Experts} method with the modern \RV{detection} and ensemble methods. For evaluation, we use PASCAL VOC (2007\&2012) \cite{pascalvoc} and ILSVRC datasets \cite{ILSVRC15}. For more evaluation of our \RV{Rank of Experts}, we participated in  ILSVRC 2017 detection competition. We use the common  metrics consisting of AP (averaged over IoU thresholds) and mAP (averaged over classes). When evaluating our methods on each set, we do not use other additional dataset (\eg. MSCOCO) for training. 

%

\textbf{Training parameters:} 
All the parameters of implemented detectors have been determined experimentally and remained unchanged for each dataset evaluation. The most parameters are same to those given in official codes of FRCN \cite{RenHGS15_NIPS15} and SSD \cite{LiuAESRFB_ECCV16}. From extensive evaluation, we find that most parameters do not affect the overall performance significantly. We set the learning rates to $10^{-3}$ and $10^{-4}$ for PASCAL VOC and ILSVRC datasets, respectively. However, we found that the fraction $FG$  of labeled foreground samples in a mini-batch and threshold $TR$, which limits the sizes of proposal boxes, \RE{can affect} the performance.  Therefore, both parameters are determined from evaluations as discussed in the next section.  \RV{Since the explicit validation sets are available in both datasets}, we evaluate \RV{the ranking matrix} with the \RV{provided} validation sets \RV{that are not exploited for training a detector.}





\begin{table}[tbp] \vspace{-5pt}
\caption{{Performance comparison with other detectors on the PASCAL VOC07/12 \RE{test sets}. The more results  can be found in our supplementary material and  \href{http://host.robots.ox.ac.uk:8080/leaderboard/displaylb.php?challengeid=11&compid=4}{the PASCAL VOC 2012 website}.}} 
 \vspace{-5pt}
{\footnotesize
\begin{center}
\renewcommand{\tabcolsep}{0.9mm}
\begin{tabular}{ccccccc}
\hline \hline
\textbf{Contributors}& \textbf{Feature} &   \textbf{Meta} & \textbf{RoI } &\textbf{mAP} &\textbf{Speed}\\
\textbf{}& \textbf{extractor} &   \textbf{architecture} & \textbf{warping} &&\textbf{(fps)}\\
\hline \hline
\multicolumn{6}{c} {Training set: VOC07 trainval / Test set: VOC07 test}\\
\cite{Girshick15_ICCV15}& VGG16 &Fast  & Pooling & 66.9\% & -\\
\cite{RenHGS15_NIPS15}& VGG16& FRCN   & Pooling & 69.9\% & -\\
\cite{LiuAESRFB_ECCV16}& VGG16 &SSD300  & - & 68.0\%  & -\\
\cite{LiuAESRFB_ECCV16}& VGG16 &SSD512  & -&71.6\% & -\\
\cite{RedmonDGF_CVPR16}&VGG16 &YOLO &-&57.9\% \\
Ours & Res269& FRCN-Type1 & Pooling &77.4\%& 3.0\\
Ours & Res101& FRCN-Type2 & Alignment &75.8\% & 6.5\\
Ours & Res152& \RV{FRCN-Type1} & Alignment & \RV{76.1}\% & 5.2\\
Ours & Res152& FRCN-Type2 & Alignment &76.2\% & 5.6\\
Ours & WRI & SSD & - & 77.0\% & 12.0 \\
Ours & VGG & SSD & - & 74.2\% & 14.8\\
Ours & VGG& DSSD & - &74.5\% & 7.6\\
\multicolumn{4}{c} {\cellcolor[gray]{0.875}\color{black}\textbf{Rank of Experts (for 7 detectors)}} &\cellcolor[gray]{0.875}\color{black}\textbf{81.2\%} & \cellcolor[gray]{0.875}\color{black}\textbf{0.87} \\
\hline \hline
\multicolumn{6}{c} {Training set: VOC07/12trainval / Test set: VOC12 test}\\
\cite{FuLRTB_Corr17}&Res101 &DSSD513 &-&80.0\% & - \\
\cite{LeeCJK_corr17}&VGG16 &RUN-3WAY &-&79.8\% & - \\
\cite{LiuAESRFB_ECCV16}& VGG16 &SSD512  & -&	78.5\% & - \\
\cite{RenHGS15_NIPS15}& Res101 &FRCN  & Pooling &	76.8\%& - \\
Ours & Res269& FRCN-Type1 & Pooling &79.3\% &3.0 \\
Ours & Res101& FRCN-Type2 & Alignment &76.9\% & 6.5\\
Ours & Res152& \RV{FRCN-Type1} & Alignment &77.9\% & 5.2\\
Ours & Res152& FRCN-Type2 & Alignment &77.2\% & 5.6\\
Ours & WRI&SSD& -&78.4\% & 12.0\\
Ours & VGG&SSD& -&76.6\% & 14.8 \\
Ours & VGG&DSSD& -&77.6\%& 7.6 \\
\multicolumn{4}{c} {\cellcolor[gray]{0.875}\color{black}\textbf{Rank of Experts (for 7 detectors)}} &\cellcolor[gray]{0.875}\color{black}\textbf{82.2\%} &\cellcolor[gray]{0.875}\color{black}\textbf{0.87}\\

\hline \hline
\end{tabular}\vspace{-10pt}
\end{center}}
\label{TABLE:Comp_PASCAL}
\end{table}

\begin{table}[tbp]
\vspace{-5pt}
\caption{{ILSVRC 2017 competition results. More detection results can be found \href{http://image-net.org/challenges/LSVRC/2017/results}{ in the ILSVRC 2017 website.}}}  
\vspace{-5pt}
{\footnotesize
\begin{center}
\renewcommand{\tabcolsep}{0.7mm}
\begin{tabular}{c|c|c}
\hline \hline
\textbf{Team name} & $\textbf{Categories Won}$ & $\textbf{mAP}$ \\
\hline 
BDAT & 85 & 73.13\% \\ 
NUS-Qihoo-DPNs  & 9 & 65.69\% \\ 
KAISTNIA-ETRI  & 0 & 61.02\% \\ 
QINIU-ATLAB+CAS-SAR  & 0 & 52.66\% \\ 
FACEALL-BUPT  & 0 & 49.58\% \\ 
PaulForDream  & 0 & 49.42\% \\ 
\cellcolor[gray]{0.875}\color{black}\textbf{DeepView (Rank of Experts)} & \cellcolor[gray]{0.875}\color{black}\textbf{10} & \cellcolor[gray]{0.875}\color{black}\textbf{59.30\%} \\ 

\hline \hline 

\end{tabular}
\end{center}}
\label{TABLE:ILSVRC_Competition}
\vspace{-10pt}
\end{table}

\begin{table*}[tbp] \vspace{-10pt}
\caption{{Evaluation results of different single and ensemble detectors on the  ILSVRC 2017 val2 set. For each image resolution, {\color{red}\textbf{the best detection rates are marked with red color}} and {\color{blue}\textbf{the number of the most selected detectors are marked with blue color.}}}} 
 \vspace{-5pt}
{\footnotesize
\begin{center}
\renewcommand{\tabcolsep}{1.2mm}
\begin{tabular}{c|c|c|c|c|ccccc}
\hline \hline

\textbf{Network}&\textbf{Feature} &   \textbf{Meta} & \textbf{RoI} &\textbf{Training} & \multicolumn{5}{c} {\textbf{mAP (\# of selection) per image resolution}}  \\
&\textbf{extractor} & \textbf{architecture} &   \textbf{warping} & \textbf{Dataset}&  \textbf{400}&  \textbf{512}&  \textbf{600}&  \textbf{800}& \textbf{900}\\
\hline \hline
\textbf{D1} &Res101 &FRCN-Type1 & Pooling &train &50.07\%  (0) &-& 53.13\% (13) &53.25\% (16) &52.08\% (10)  \\
\textbf{D2}  &	Res101&	FRCN-Type1&	Pooling&	train+val1&	49.79\% (2)  & - & 53.57\% (12) & 53.32\% (7)  &51.96\% (5) \\
\textbf{D3}  &	Res152&	FRCN-Type1&	Pooling&	train+val1&	52.16\% (11) & -	&	55.77\% (32) & 54.94\% (20)& 53.59\% (19)  \\
\textbf{D4}  &	Res152&	FRCN-Type1&	Pooling&	train&	52.59\% (13) & - &	55.71\% (23) &	55.27\% (26)  & - \\
\textbf{D5}  &	Res152&	FRCN-Type1&	Pooling&	train+val1&	51.41\% (6) & -&	55.35\% (32) &	54.21\% (19) & 52.91\% (11)  \\
\textbf{D6}  &	Res152&	FRCN-Type1&	Pooling&	train&	51.92\% (5)  & - &	56.00\% (28) &	55.41\% (18) & 54.38\% (24)  \\
\textbf{D7}  &	Res152&	FRCN-Type1&	Pooling&	train+val1&	52.41\% (8)  &	-& 56.19\% (32)  & 55.54\% (25)  & 54.34\% (22)   \\
\textbf{D8}  &	Res269&	FRCN-Type1&	Pooling&	train+val1&	-& - & 56.92\% (49) & - & - \\
\textbf{D9}  &	Res269&	FRCN-Type1&	Pooling&	train+val1&	54.09\% \color{blue}\textbf{(24)}& -	&  57.65\% (69)&	56.29\% (38)&	54.94\% (19) \\
\textbf{D10}  &	Res269&	FRCN-Type1&	Pooling&	trainval+val1+aug&	53.21\% (17)& - &56.76\% (43)&	55.84\% (34)&	54.49\% (20) \\
\textbf{D11}  &	Res269&	FRCN-Type1&	Pooling&	trainval1&\color{red}{\textbf{53.98\%}} \color{black}(23)& - &\color{red}\color{red}\textbf{57.72\%} \color{blue}\textbf{(76)}&\color{red}\color{red}\textbf{56.64\%} \color{blue}\textbf{(45)}&	55.41\% (26) \\
\textbf{D12}  &	Res269&	FRCN-Type1&	Pooling&	train&	53.59\% (17)& - & 57.34\% (63)&	56.56\% (33)&\color{red}\color{red}\textbf{55.53\%}\color{black}(29) \\
\textbf{D13}  &	Res152&	FRCN-Type2&	 Alingment&	train+val1&	48.91\% (6)& - & 54.65\% (46)&	54.53\% (30)&	54.49\% \color{blue}\textbf{(42)} \\
\textbf{D14} &	Res152&	FRCN-Type2&	 Alingment &trainval+val1+aug & - & - &54.57\% (40)& 53.92\% (29)&	- \\
\textbf{D15}  &	Res152 &	FRCN-Type2&	 Alignment&	train+val1&	51.95\% (8)& - &56.15\% (35)&	55.41\% (25)&	54.73\%  (20) \\
\textbf{D16}  &	Res152 &	FRCN-Type2&	 Alignment&	trainval+val1+aug&	43.42\% (1)& - &51.39\% (10)&	49.00\% (5)& 47.40\% (2) \\	
\textbf{D17} &	VGG&	SSD&	- &	trainval+val1+aug & -&50.48\% (14) & - & -& - \\			
\textbf{D18} &	VGG&	DSSD &	-	&trainval+val1+aug& -&		49.98\% (15) & - & -& - \\			
\textbf{D19} &	WRI&	SSD&	- &	trainval+val1+aug & - &49.21\% (8) & - & -& -\\	
\hline
\multicolumn{5}{c} {\cellcolor[gray]{0.875}\textbf{Rank of Experts (for 19 detectors)}}  &\multicolumn{5}{c}{\cellcolor[gray]{0.875}\color{red}\textbf{62.54\%}} \\		
\hline \hline
\end{tabular}\vspace{-10pt}
\end{center}}
\label{TABLE:Comp_ILSVRC}
\end{table*}

\begin{figure*}[!tbp]
\begin{center}
\vspace{-8pt}
\includegraphics[width=0.24\linewidth, height=2.4cm]{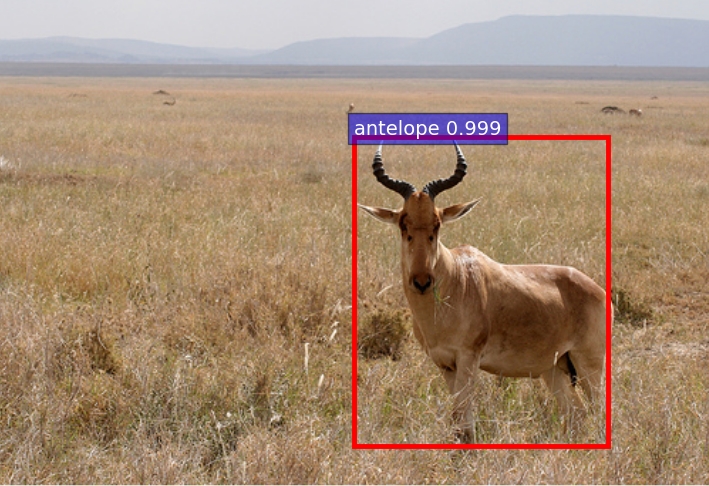}
\includegraphics[width=0.24\linewidth, height=2.4cm]{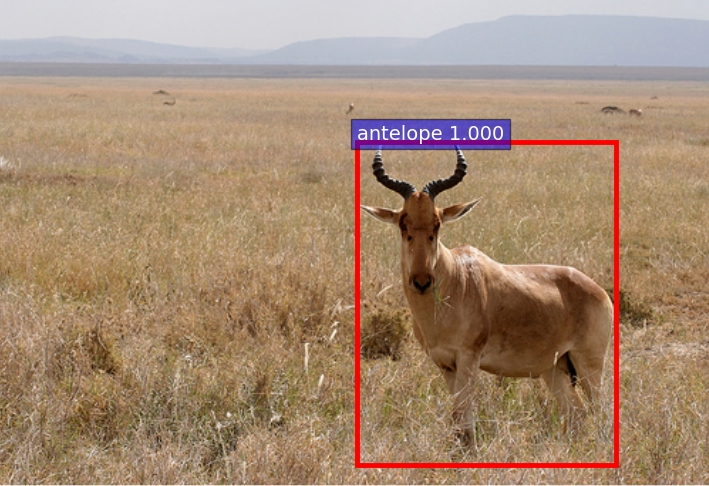}
\includegraphics[width=0.24\linewidth, height=2.4cm]{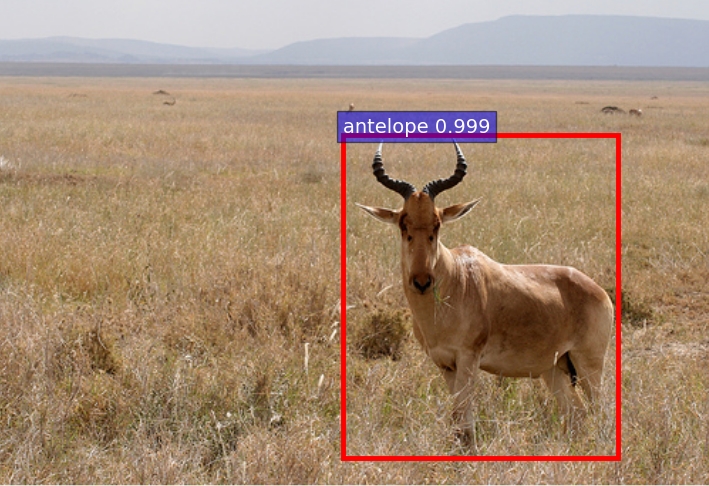}
\includegraphics[width=0.24\linewidth, height=2.4cm]{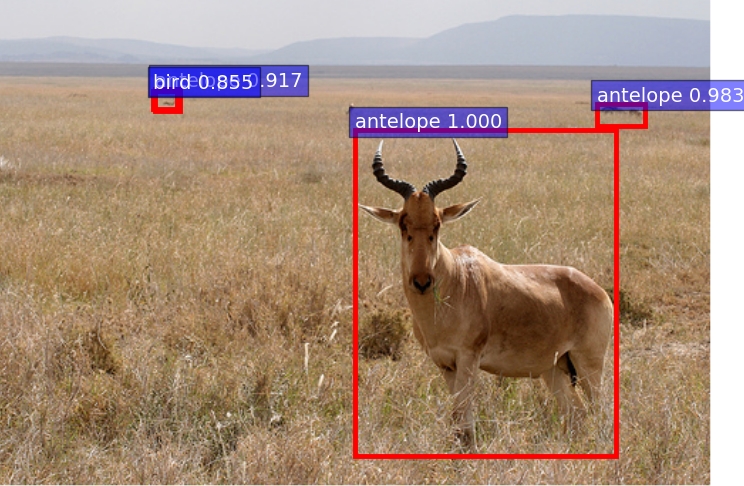}
\\
\includegraphics[width=0.24\linewidth, height=2.4cm]{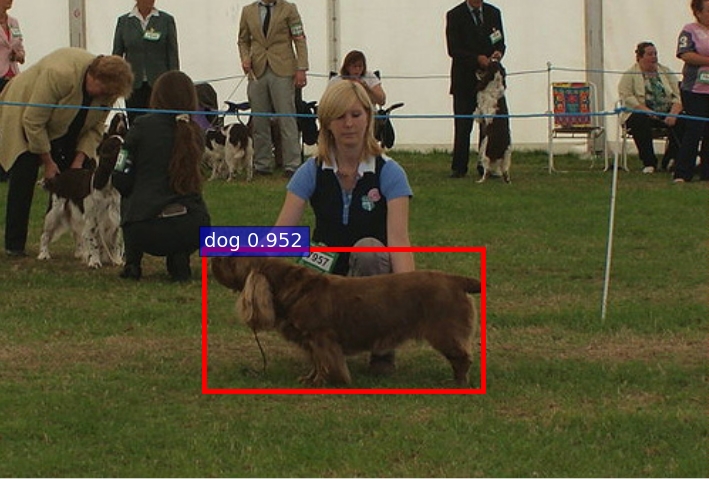}
\includegraphics[width=0.24\linewidth, height=2.4cm]{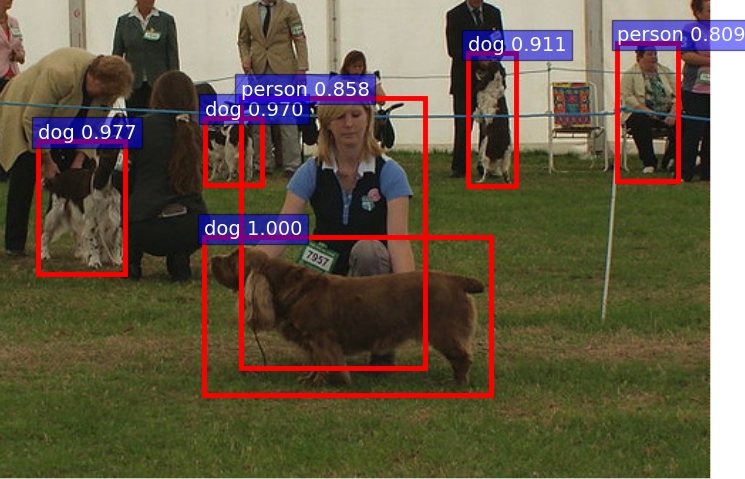}
\includegraphics[width=0.24\linewidth, height=2.4cm]{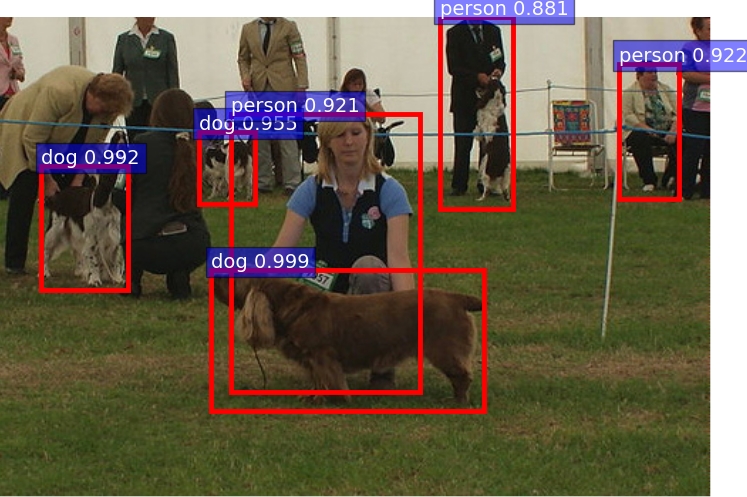}
\includegraphics[width=0.24\linewidth, height=2.4cm]{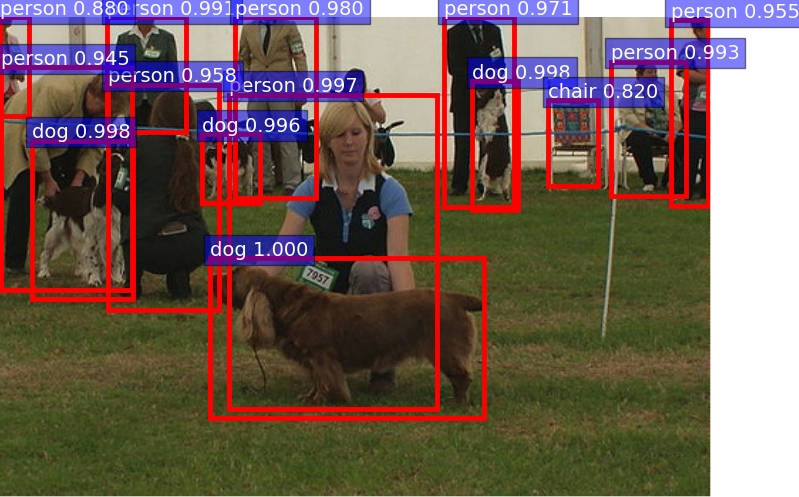}
\\
\includegraphics[width=0.24\linewidth, height=2.4cm]{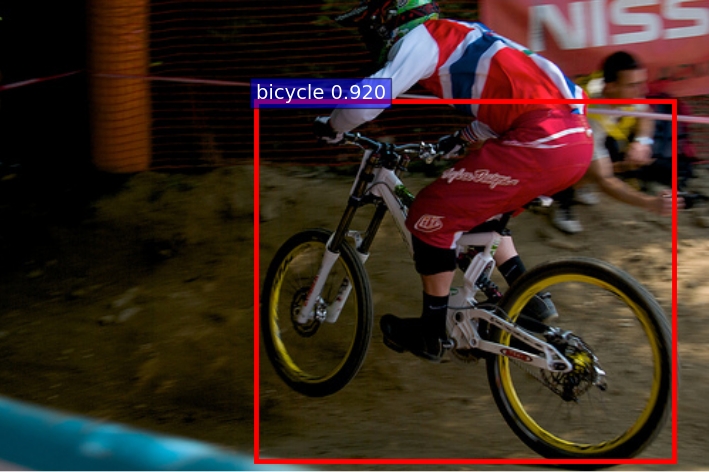}
\includegraphics[width=0.24\linewidth, height=2.4cm]{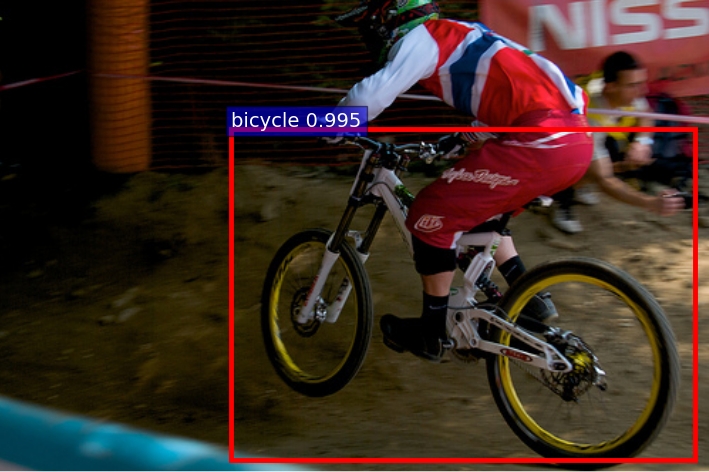}
\includegraphics[width=0.24\linewidth, height=2.4cm]{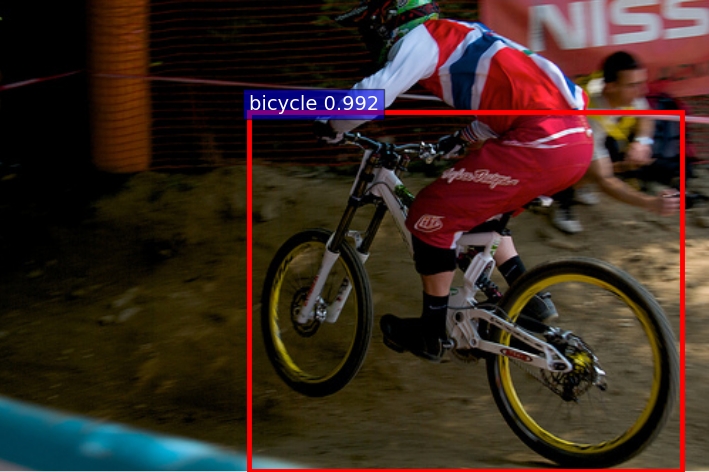}
\includegraphics[width=0.24\linewidth, height=2.4cm]{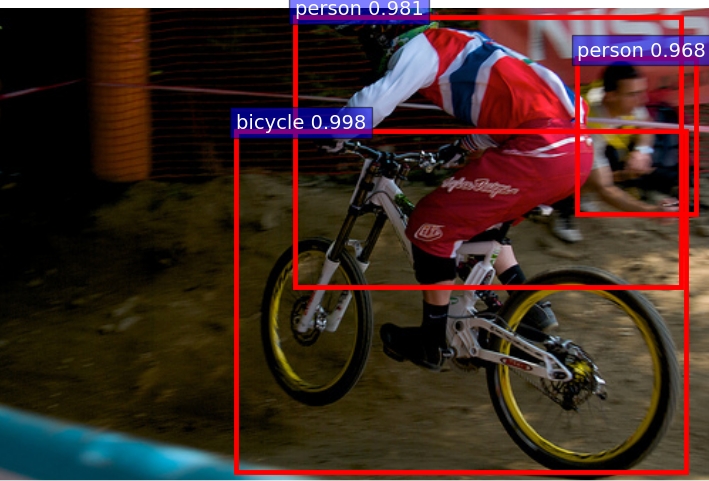}
\\
\includegraphics[width=0.24\linewidth, height=2.4cm]{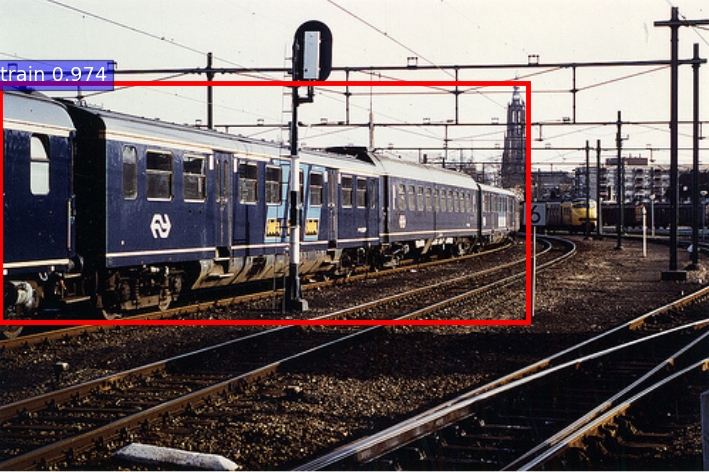}
\includegraphics[width=0.24\linewidth, height=2.4cm]{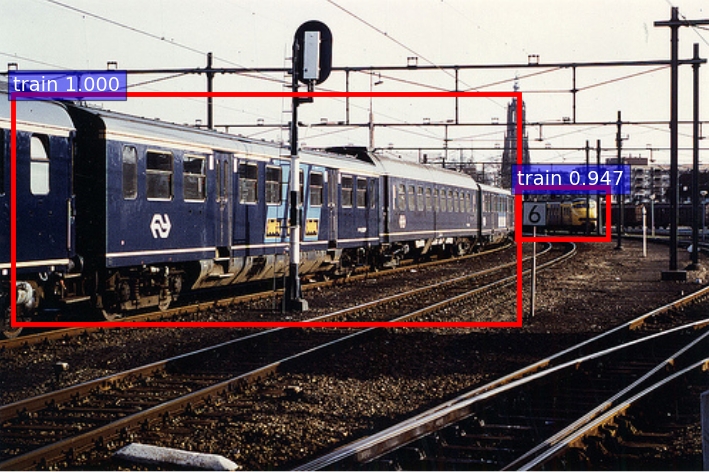}
\includegraphics[width=0.24\linewidth, height=2.4cm]{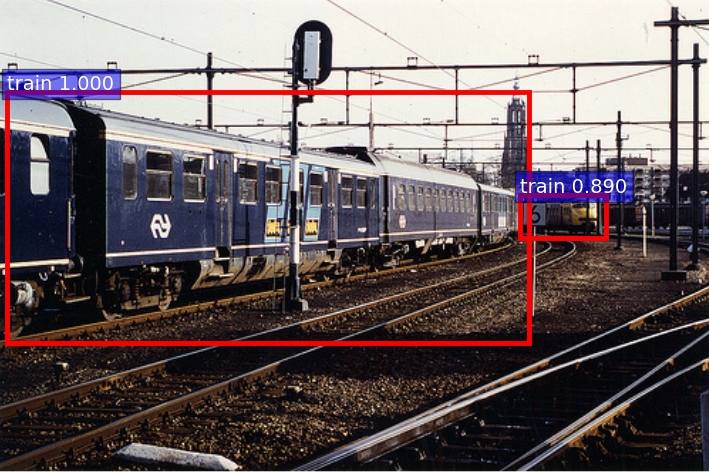}
\includegraphics[width=0.24\linewidth, height=2.4cm]{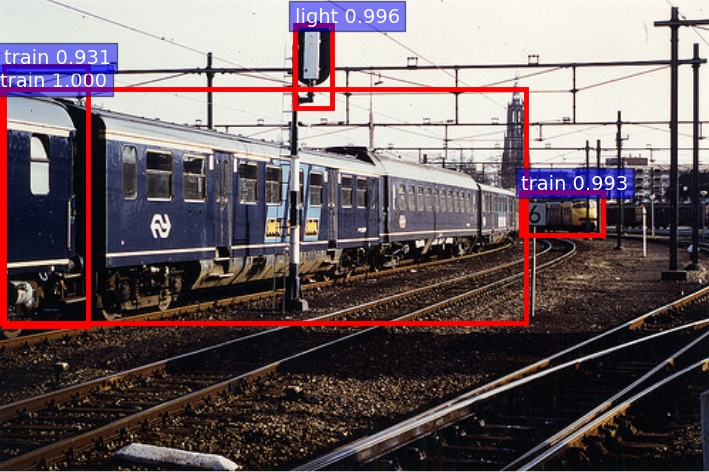}
\\
\vspace{-5pt}
\end{center}
\caption{{Detection results of VGG-DSSD, Res152-FRCN, Res269-FRCN and Rank of Experts are shown in each column.}}
\vspace{-5pt}
\label{fig:Object_detection}
\end{figure*}


\begin{figure*}[!tbp]
\begin{center}
\vspace{-10pt}
\subfigure[PASCAL 2007 test set]{\includegraphics[width=0.40\linewidth]{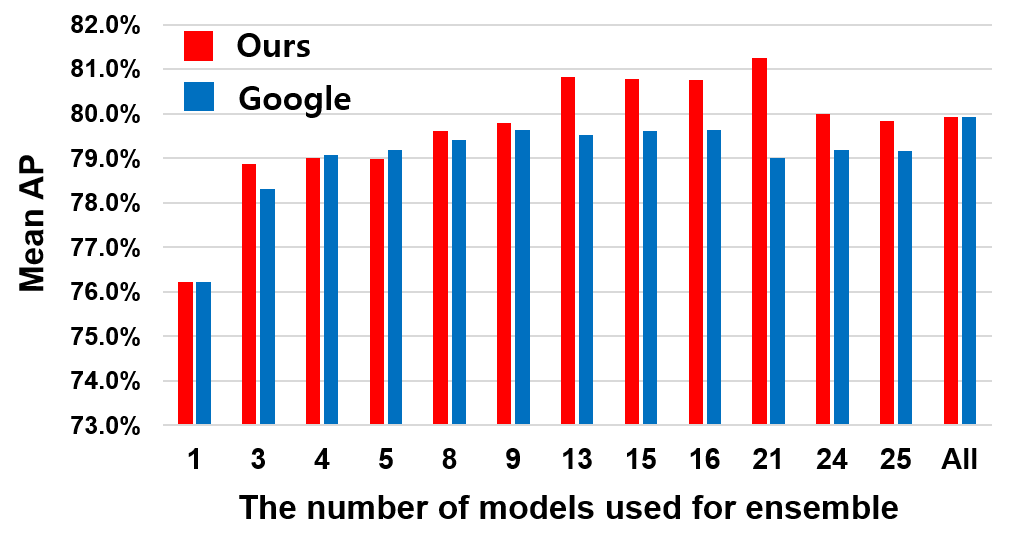}}
\subfigure[ILSVRC 2017 val2 set]{\includegraphics[width=0.40\linewidth]{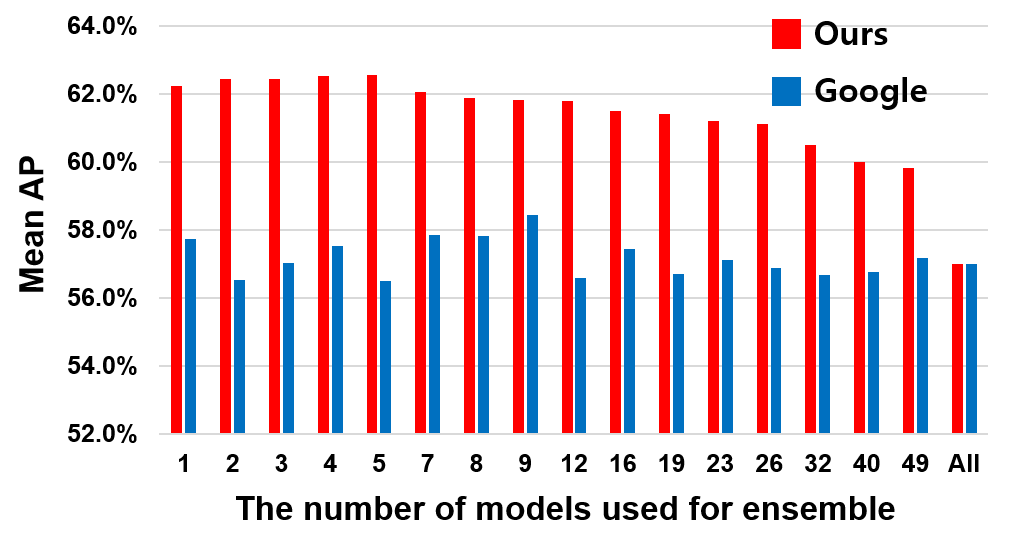}}
\end{center}
\vspace{-5pt}
\caption{\RV{Performance comparison with other ensemble method \cite{HuangRSZKFFWSG_Corr16} on PASCAL and ILSVRC datasets.}}
\vspace{-5pt}
\label{fig:Ens_comp}
\end{figure*}

\textbf{Implemented detectors:} 
As shown in Table  \ref{TABLE:Comp_ILSVRC}, we implemented   19 detectors with different feature extractors, meta-architectures, and training sets. All the details of those can be found in \RV{Sec.} \ref{Sec:Pool}.  In  \RV{Table} \ref{TABLE:Comp_PASCAL}-\ref{TABLE:ILSVRC_Ab}, the performance of our detectors are highlighted.

\textbf{PASCAL VOC evaluation:} 
For VOC 2007 and 2012 (VOC07/12) tests, we train detectors with VOC07trainval (5011 images) and VOC07/VOC12trainval (21503 images) sets as also done in \cite{Girshick15_ICCV15, RenHGS15_NIPS15, LiuAESRFB_ECCV16, RedmonDGF_CVPR16}. We train detectors with \RV{$\mu=10^{-3}$} learning rate for 50k iterations, and continue the training for 30k iteration with \RV{$\mu=10^{-4}$} for VOC07. For VOC12, we  train the detectors during 70k and 50k iterations with \RV{the same learning rates}. 

In Table \ref{TABLE:Comp_PASCAL}, we compare the implemented detectors with the modern convolutional detectors. For the VOC07 test set, all the implemented detectors show the better results than the original ones. Main reason is that we more successfully train detectors by using \RV{feature aggregation} in Sec. \ref{Sec:Meta} and tunning important parameters well \RV{as} shown in  \RV{Table} \ref{TABLE:Comp_Param}. In particular, we obtain the best detection rates using \RV{the} Res269-FRCN models. Exploiting feature extractors with low classification errors  also improves the detection rates when comparing Res101/Res152/Res269-FRCN models.  \RV{In addition, SSD and DSSD detectors show the comparable performance with the FRCN detectors.  We can improve the detection rate about 4\% using ensemble of 7 detectors.}


We further evaluate our methods in the PASCAL VOC 2012 challenge. In this evaluation, we only use the VOC dataset (without using COCO dataset) for a fair comparison. The comparison results are given in \RV{Table} \ref{TABLE:Comp_PASCAL}. Compared to the state-of-the-art detectors, the implemented detectors \RV{show} the competitive performance, and we achieve the best mAP \RV{by} using \RV{the proposed ensemble detection}.

\textbf{ILSVRC 2017 competition:}
For this competition, we implement 19 detectors with different structures, training sets and \RV{initial states} as shown in Table \ref{TABLE:Comp_ILSVRC}. \RV{We train the detectors with the $\mu = 10^{-4}$ and $\mu = 10^{-5}$ for 1M and 400k iterations}. We also provide the performance of each detector. Among them D11 using RES269-FRCN produces the \RV{best} mAP. Moreover, the mAP difference between detectors is much \RV{bigger} than on PACAL VOC evaluation. The more \RV{comparison} results are given in Fig. \ref{fig:Object_detection}. Our Rank of Experts shows the better results than other single detectors.

%

Table \ref{TABLE:ILSVRC_Competition} shows the comparison results with detectors of other teams in ILSVRC 2017 competition.  Although three teams \RV{show the better} mAPs than \RV{ours}, we won the 2nd place for object category won. These results prove that the proposed Rank of Expert \RV{can} improve class-wise detection \RV{rates significantly}.

\begin{table}[tbp]
\caption{Mean AP(mAP) improvement on ILSVRC 2017 val2 set.}  
\vspace{-5pt}
{\footnotesize
\begin{center}
\renewcommand{\tabcolsep}{0.7mm}
\begin{tabular}{cc|cc}
\hline \hline
\textbf{Method} & $\textbf{mAP}$ & $\textbf{Method}$& $\textbf{mAP}$ \\
\hline 
Soft-NMS \cite{BodlaSCD_CORR17} & $\sim 1\%$ &Data augmentation  & $1 \sim 2 \%$ \\
Multi-scale test & $\sim 1\%$ &\cellcolor[gray]{0.875}\color{black}\textbf{Rank of experts} & \cellcolor[gray]{0.875}\color{black}\textbf{4} \textbf{$\sim$} \textbf{5 \%} \\
\hline \hline 
\end{tabular}
\end{center}}
\vspace{-10pt}
\label{TABLE:ILSVRC_Ab}
\end{table}

\textbf{Ablation experiment:}
In Table \ref{TABLE:ILSVRC_Ab}, we show the performance improvement when applying each method. We evaluate the averaged mAP improvement of D10, D14 and D16-19 detectors. We found that the mAP is \RV{greatly} enhanced  by using the proposed Rank of Experts.

\textbf{Comparison with other ensemble method:}
To show effectiveness of our \RV{Rank of Experts}, we compare our method with the ensemble method of Google \cite{HuangRSZKFFWSG_Corr16} which won 1st places on  \RV{the MSCOCO 2016 competition}. To this end, we evaluate a similarity matrix by computing cosine distances between the AP score vectors of all the detectors. We then greedily select the best models with highest mAP, but prune away a model when the similarity score between selected ones and the candidate exceeds to a threshold. By changing the threshold, we can vary the number of detectors used for ensemble.

Figure \ref{fig:Ens_comp} shows the comparison results of both ensembe methods on PASCAL VOC 2007 and ILSVRC 2017 datasets. In the most cases, we achieve the better mAPs on both datasets. On average the proposed method improves the mAPs to $0.31\%$ and $4.15\%$ for PASCAL and ILSVRC \RV{datasets}, respectively. This indicates that our method is more effective especially for large \RV{scale} object detection.

\textbf{Discussion:} 
\RV{For our Rank of Experts, the most important parameter is $\delta= [0,1]$ that determines $N_{j}$ of how many detectors will be used for each class detection}.  If $\delta=0$, we can use one detector with the best mAP only, but use all the detectors if $\delta=1$. The more detectors \RV{can be used for ensemble} when $\delta$ increases.

To show the effects of $\delta$, we change the $\delta$ from 0 to 0.1 with interval 0.05 on PASCAL and ILSVRC datasets. For this evaluation, 7 detectors (PASCAL) and 19 detectors (ILSVRC) with multi-scale testing as described in Table \ref{TABLE:Comp_PASCAL} and \ref{TABLE:Comp_ILSVRC} were used. We obtain the best mAP $81.24\%$ and $62.54\%$ with $\delta=0.03$ for each dataset. However, the number of detectors used for ensemble on each dataset is different. More detectors were used on PASCAL evaluation \RV{since the mAP difference of detectors is not significant}. \RV{Note that the performance difference of ensemble detection  using different $\delta$ is negligible.} There are only  $\sim 1\%$ and $\sim 2\%$ mAP differences for the PASCAL and ILSVRC datasets. \RV{Thus, we verify that $\delta$ does not affects the overall performance significantly and the robustness of our \RV{Rank of Experts}.}

In Table \ref{TABLE:Comp_ILSVRC}, we provide the number of selected detectors \RV{for ensemble}. In general, detectors with high mAPs are selected more frequently. However, \RV{all the detectors} can be used for ensemble detection \RV{by using the proposed Rank of Experts} although some detectors (\ie. WRI-SSD and VGG-DSSD) have much lower mAPs than other \RV{detectors}. 

\textbf{Running time:}
The speed of each detector depends on types of feature extractors and meta-architectures. We present the speed of our implemented detectors in Table \ref{TABLE:Comp_PASCAL}. The running time of our Rank of Experts relies on the number of detection boxes. \RV{For 162K and 3.66M boxes}\footnote{For PASCAL dataset, 27 (models) $\times$ 20 (classes) $\times$ 300 (maximum number of boxes per class). For ILSVRC dataset, 61$\times$ 200  $\times$ 300.} on PASCAL and ILSVRC dataset evaluation, the speeds are $\sim$ ~140(fps) and  $\sim$ 9.37(fps), respectively. The main bottleneck occurs during Soft-NMS. The speed can be greatly improved with the GPU programming.

\section{Conclusion}
In the recent object detection challenges, the top-ranked teams significantly improve the detection rates using \RE{ensemble learning}. However, the conventional methods requires additional training stages with huge GPU memory \RE{usage}. In this paper, we therefore propose an effective ensemble \RE{detection} for combining different detectors \RE{without the expensive training}. Furthermore, the additional benefit of \RE{our ensemble detection} can improve the class-wise detection rate directly. In order to increase the diversity between detectors, we implement a variety of detectors with different feature extractors, meta-architectures and training sets. We have proved that the proposed method leads to the obvious performance improvement through extensive evaluations.

{\small
\bibliographystyle{ieee}
\bibliography{Rank_bsh}
}

\end{document}